\documentclass{article}

\usepackage{microtype}
\usepackage{graphicx}
\usepackage{subfigure}
\usepackage{booktabs} 

\usepackage{hyperref}

\usepackage[accepted]{icml2024}

\usepackage{amsmath}

\usepackage{amssymb}
\usepackage{mathtools}
\usepackage{makecell}
\usepackage{amsthm}

\usepackage[capitalize,noabbrev]{cleveref}

\theoremstyle{plain}
\newtheorem{theorem}{Theorem}[section]

\theoremstyle{definition}
\newtheorem{definition}[theorem]{Definition}

\theoremstyle{remark}

\usepackage[textsize=tiny]{todonotes}

\icmltitlerunning{Submission and Formatting Instructions for a conference}

\begin{document}

\twocolumn[
\icmltitle{SAR-AE-SFP: SAR Imagery Adversarial Example in Real Physics domain with Target Scattering Feature Parameters}



\icmlsetsymbol{equal}{*}

\begin{icmlauthorlist}
\icmlauthor{Jiahao Cui}{sch}
\icmlauthor{Jiale Duan}{sch}
\icmlauthor{Binyan Luo}{sch}
\icmlauthor{Hang Cao}{sch}
\icmlauthor{Wang Guo}{sch}
\icmlauthor{Haifeng Li}{sch}

\end{icmlauthorlist}

\icmlaffiliation{sch}{School of Geosciences and Info-Physics, Central South University, Changsha 410083, China}

\icmlcorrespondingauthor{Haifeng Li}{lihafeng@csu.edu.cn}

\icmlkeywords{Machine Learning, ICML}

\vskip 0.3in
]



\printAffiliationsAndNotice{}  

\begin{abstract}
Deep neural network-based Synthetic Aperture Radar (SAR) target recognition models are susceptible to adversarial examples. Current adversarial example generation methods for SAR imagery primarily operate in the 2D digital domain, known as image adversarial examples. Recent work, while considering SAR imaging scatter mechanisms, fails to account for the actual imaging process, rendering attacks in the three-dimensional physical domain infeasible, termed pseudo physics adversarial examples. To address these challenges, this paper proposes SAR-AE-SFP-Attack, a method to generate real physics adversarial examples by altering the scattering feature parameters of target objects. Specifically, we iteratively optimize the coherent energy accumulation of the target echo by perturbing the reflection coefficient and scattering coefficient in the scattering feature parameters of the three-dimensional target object, and obtain the adversarial example after echo signal processing and imaging processing in the RaySAR simulator. Experimental results show that compared to digital adversarial attack methods, SAR-AE-SFP Attack significantly improves attack efficiency on CNN-based models (over 30\%) and Transformer-based models (over 13\%), demonstrating significant transferability of attack effects across different models and perspectives.
\end{abstract}

\section{Introduction}
\label{submission}

Deep neural network (DNN)-based Synthetic Aperture Radar (SAR) image target recognition algorithms have the advantage of end-to-end feature learning, which can effectively improve target recognition rates \cite{chen2014sar}. However, studies show that DNN models are susceptible to interference from adversarial examples. Attackers can cause these models to produce incorrect recognition results by adding meticulously designed minor perturbations to SAR images \cite{szegedy2013intriguing}. The presence of adversarial examples poses a serious threat to SAR target recognition tasks. Hence, researching adversarial examples in SAR imagery is crucial for enhancing the security and robustness of SAR systems and preventing potential risks from adversarial attacks.

Existing SAR image adversarial example generation techniques can be divided into two categories: image adversarial examples and pseudo physics adversarial examples. Image adversarial examples are created by adding subtle perturbations to time or frequency domain images \cite{zhang2022generating}, which are hard to detect by the human eye, through methods such as gradient optimization \cite{li2020adversarial,huang2020adversarial1,inkawhich2020advanced}, constraint optimization \cite{sun2021adversarial,du2021fast}, estimating decision boundaries \cite{huang2020adversarial2}, and using Generative Adversarial Networks (GAN) \cite{wang2021adversarial,du2021adversarial}. Pseudo physics adversarial examples, on the other hand, are generated through either constructing attribute scattering center models of adversarial scatterers \cite{dang2021sar,peng2022scattering}   or employing phase modulation methods \cite{liu2021sar,xia2022sar}. These approaches produce adversarial perturbations with clear physical significance on SAR temporal domain images, specifically aimed at executing adversarial attacks against target classifiers.

However, the pixel values in SAR images represent the coherent energy accumulation of echo signals \cite{cummingf2004digital}, and transferring or reproducing optical image adversarial attack methods has not established a correspondence between perturbed pixels and radar signals. Pseudo physics adversarial examples are realized based on clear physical electromagnetic scattering features and phase information \cite{jackson2006feature}, giving adversarial perturbations in the image domain a better physical realization prospect. They still use static single-frame images as the basis for attacks, failing to model the perturbation changes during the target's motion process. Therefore, the aforementioned attack methods struggle to generate real physics adversarial examples.

As mentioned above, the pixel values in SAR images reflect the target object's reflection strength towards radar beams, including amplitude and phase information, which carry rich details about the target object such as its geometric shape and scattering properties. Therefore, intuitively, by altering the scattering feature parameters of the ground target, one can change the intensity of the echo signal, thereby generating real physics adversarial examples.

\begin{figure}[!t]
\vskip 0in
\centerline{\includegraphics[width=\columnwidth]{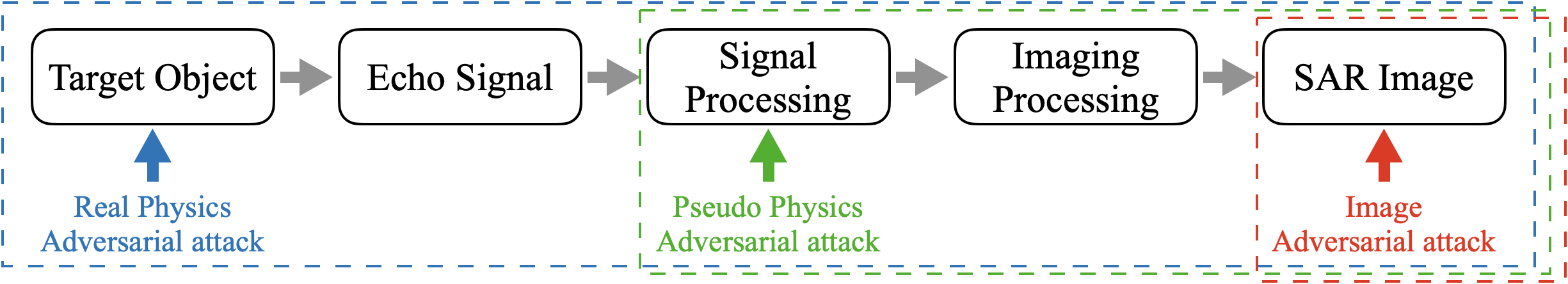}}
\caption{The positions where different attack methods add perturbations in the imaging chain.}
\label{figure1}
\vskip -0.1in
\end{figure}

Based on this, we propose SAR-AE-SFP-Attack, a method to generate real physics adversarial examples by changing the scattering feature parameters of target objects. Compared to existing methods, the SAR-AE-SFP-Attack method adds perturbations directly to the target object, achieving a full-chain attack in the SAR imaging process (Fig. 1). Specifically, we first model the adversarial attack as changing the reflection and scattering coefficients in the target's scattering feature parameters. Then, using the RaySAR physical simulator \cite{auer2016raysar}, we simulate the scattering process of electromagnetic waves on the target, allowing for real-time or near real-time simulation of the target's echo signal amplitude under any posture, configuration, or observation geometry. After signal processing and imaging processing, we obtain SAR adversarial examples. Finally, we develop an innovative finite difference method to adapt to the non-differentiable nature of the RaySAR simulator. This method models the generation of adversarial examples as an optimization problem, effectively estimating gradients by perturbing the reflection and scattering coefficients in the target's scattering feature parameters slightly and efficiently updating these parameters using batch processing techniques. Adversarial examples generated in this way enable the target object to maintain adversarial characteristics in multi-perspective physical scenarios.

\section{Related work}

Adversarial examples were first introduced by \cite{goodfellow2014explaining}, they proved that by adding meticulously designed minor perturbations $\delta$ to SAR images $x$, a well-trained deep neural network model $f$ can be deceived into producing incorrect recognition results. The formulation is expressed as:
\begin{equation}\label{eq1}
	f(x+\delta) \neq f(x), s.t. \parallel \delta  \parallel \le \epsilon
\end{equation}
In this context, $\parallel \delta  \parallel$ represents the size of the perturbation, and $\epsilon$ denotes the maximum allowable value of the perturbation. We classify existing work into two categories: image adversarial attacks and pseudo physics adversarial attacks, based on whether they integrate the imaging scatter mechanism of SAR.

\subsection{Image Adversarial Attack}

The method of generating image adversarial examples draws inspiration from adversarial attack methods in optical images, creating adversarial perturbations on a per-pixel basis. This type of method has been extensively studied in the initial stages of adversarial attacks on SAR images, typically involving empirical exploration and evidential research. Unlike optical image target recognition, the input data for SAR target recognition systems can be either temporal domain grayscale images or frequency domain complex images. Therefore, image adversarial attacks can be further divided into two forms: temporal domain image attacks and frequency domain image attacks.

\subsubsection{Temporal Domain Image Attack}

The method for generating adversarial examples for temporal domain images treats SAR images as single-channel grayscale images. By adding subtle changes to the grayscale image that are difficult for humans to recognize, deep neural network models can be misled into making incorrect classification decisions. Depending on the area of perturbation added in SAR temporal domain images, adversarial examples for SAR temporal domain grayscale images can be further divided into two types: global perturbations and local perturbations.

\textbf{Global perturbation} refers to the method of generating adversarial examples by adding perturbations across the entire input image. Currently, most global perturbations for SAR time-domain images are designed for white-box scenarios through an end-to-end approach, including methods based on gradients \cite{gao2023sar}, optimization \cite{du2021fast}, and decision boundaries \cite{wang2021universal}. However, in real-world situations, attackers typically can interact with the model but cannot access any other information about it. Therefore, black-box attack methods based on features \cite{lin2023boosting,chen2023positive}, transferability \cite{peng2022empirical} and GANs \cite{du2023tan} are widely applied in generating adversarial examples for SAR temporal domain images, effectively enhancing the transferability of adversarial examples.

\textbf{Local perturbation} refers to an attack method that adds disturbances in the target area of the input image to generate adversarial examples. Due to the obstruction of electromagnetic waves by the target during their propagation, corresponding areas in SAR images are imaged as shadow regions. For these shadow and background speckle areas, adding perturbations might violate SAR imaging characteristics and be difficult to detect manually. Inspired by sparse adversarial perturbation methods \cite{fan2020sparse,peng2022speckle}, researchers have focused on adding perturbations only to parts of the image that describe the target's reflectivity \cite{meng2022target,du2022ulan,zhang2022adversarial}. This approach effectively carries out adversarial attacks by perturbing only a quarter or less of the SAR image area.

\subsubsection{Frequency Domain Image Attack}

SAR remote sensing images are high-resolution two-dimensional frequency domain complex images \cite{chierchia2017sar,henry2018road}, containing not only amplitude information but also phase information. \cite{zhang2022generating,peng2023low} inspired by the stAdv algorithm which optimizes offset values in smooth flow fields to transform the original sample space for adversarial attacks \cite{xiao2018spatially}, proposed a new method to generate adversarial examples on SAR complex images by optimizing flow fields in the frequency domain. Compared to adversarial attack methods based on temporal domain grayscale images, attack methods based on frequency domain complex images align more closely with the data characteristics of SAR. The adversarial examples generated using these methods demonstrate improved attack effectiveness and cross-domain transferability.

\subsection{Pseudo Physics Adversarial Attack}

Pseudo physics adversarial examples are created by generating adversarial perturbations with clear physical significance on SAR temporal domain images, either through attribute scattering center models or phase modulation methods. The perturbations, produced by reconstructing attribute scattering centers \cite{zhou2023attributed,qin2023scma} or using attribute scattering center models to generate parameterized adversarial scattering centers \cite{ye2023realistic}, are added to clean examples. Furthermore, based on the ability to flexibly adjust the phase of electromagnetic wave transmission in SAR echo signals, \cite{xia2022sar,doom2023saten} combined the attack algorithm with a two-dimensional modulation interference mechanism to propose a method for generating SAR image adversarial examples aimed at real-world scenarios. However, they still use static single-frame images as the basis for attacks, failing to model the perturbation changes during the target's motion process. Therefore, the aforementioned attack methods struggle to generate real physics adversarial examples.

\section{Method}

This paper models the adversarial attack process as a precise adjustment of target scattering feature parameters and utilizes SAR image simulation technology to generate adversarial examples. This approach not only significantly reduces experimental costs and avoids uncertainties in real-world environments but also ensures that the added adversarial perturbations have realistic physical properties. The ray-tracing-based SAR image simulation method incorporates the calculation of reflectivity material properties into each mesh of the 3D model, allowing for the simulation of multiple bounces of radar signals off the target and the complex interplay of shadowing and occlusion effects between target geometrical structures, resulting in more realistic imaging outcomes. Therefore, we use the RaySAR ray tracer to simulate SAR images and generate SAR adversarial examples by perturbing the scattering feature parameters of 3D target objects.

\subsection{SAR-AE-SFP-Attack Framework}

The SAR-AE-SFP-Attack method for generating SAR adversarial examples by perturbing the scattering feature parameters of 3D target objects can be divided into two modules: the adversarial example generation module and the scattering feature parameter optimization module. The network flowchart is as follows:

\begin{figure}[ht]
\vskip 0.2in
\begin{center}
\centerline{\includegraphics[width=\columnwidth]{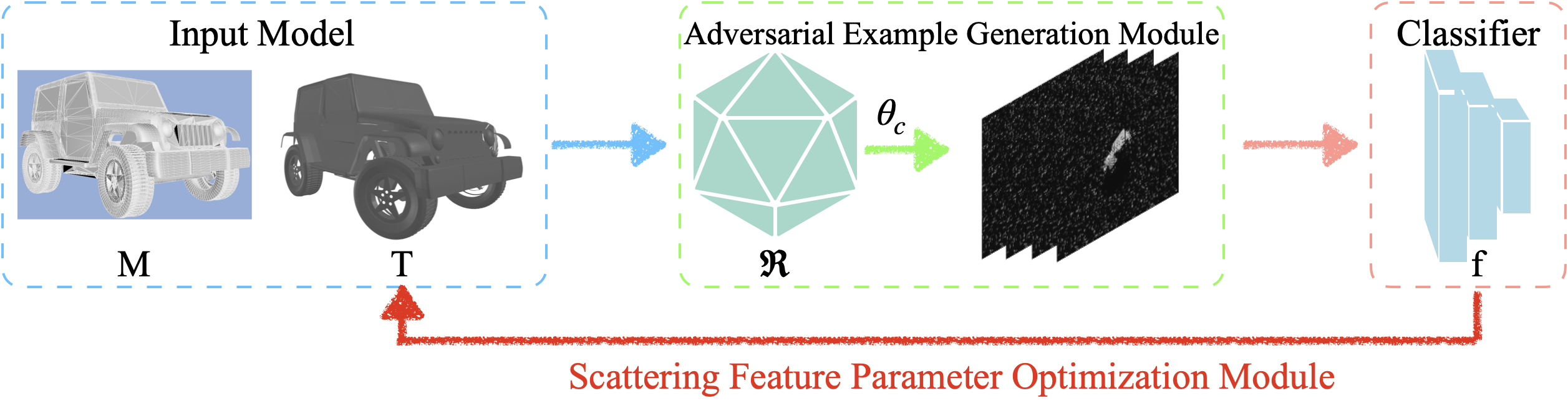}}
\caption{SAR-AE-SFP-Attack method network flowchart.}
\label{figure2}
\end{center}
\vskip -0.2in
\end{figure}

Here, $M=\{m_1,m_2,m_3,...,m_n\}$ is used to represent the geometry of the target object. Taking the JEEP model as an example, its geometry is constituted by thousands of meshes of various sizes, denoted by $m_i$; $T=\{t_1,t_2,t_3,...,t_n\}$ represents the scattering feature parameters of the meshes composing the target model. The scattering feature parameters $t_i$ include the specular reflection coefficient $F_s$, the diffuse reflection coefficient $F_d$, the surface roughness factor $F_r$, and the surface brightness factor $F_b$, among others. 
\begin{definition}
\label{Adversarial examples generation mechanism}
(Adversarial examples generation mechanism). For $t_i \in T$, the adversarial example $x'$ can be defined as \\
\begin{center}
$x'=\Re (M,T_{adv};\theta_r)$
\end{center}
where $T_{adv}=\{t_i+\delta|t_i \in T\}$.
\end{definition}
Herein, $\Re$ represents the ray tracer; $\theta_r$ is the camera parameters, representing the camera's position and its rotation angles relative to the target object. In the attack process, we introduce perturbations into the scattering feature parameters of each mesh composing the target model to obtain adversarial scattering feature parameters $T_{adv}$. Then, through SAR image simulation, adversarial examples $x'$ are obtained.
\begin{definition}
\label{Scattering feature parameters optimization mechanism}
(Scattering feature parameters optimization mechanism). For $t_i \in T$, the final adversarial scattering feature parameters $T_{adv}^*$ can be defined as \\
\begin{center}
$T_{adv}^* = \underset{T_{adv}}{\operatorname{arg\,max}} J(f(\Re(M,T_{adv};\theta_r);\theta_f),Y)$
\end{center}
where $T_{adv}=\{t_i+\delta|t_i \in T\}$.
\end{definition}
$f$ represents the  the classification model; $Y$ represents the true category of the target; $\theta_f$ signifies the parameters of the classification model. In the scattering feature parameter optimization mechanism, we iteratively optimize the adversarial scattering feature parameters $T_{adv}$ using a custom loss function $f$, resulting in the final adversarial scattering feature parameters $T_{adv}^*$ \textbf{(Definition 3.2)}. When adversarial examples containing the attribute are $T_{adv}^*$ input into the classifier $f$, they can induce the classifier to produce more misleading incorrect identification results.

\subsection{Adversarial Example Generation}

We use the RaySAR simulation software, developed and made open-source by Dr. Stefan Auer \cite{auer2016raysar}, to generate adversarial examples. The simulation process of this software can be mainly divided into two parts. Firstly, it is necessary to calculate the intensity of the echo signals received by the radar receiver after the emitted electromagnetic waves undergo multiple reflections. Secondly, the focusing position of the signal echoes in the range-azimuth plane needs to be calculated. These two parts correspond to the electromagnetic scattering model and the imaging model in the simulation method, respectively. 

In this process, the electromagnetic scattering model uses specular and diffuse reflection models in the field of optics to approximate the calculation of the echo intensity of radar signals when they are multiply bounced between objects and reflected from objects to the radar antenna. The formulas for specular and diffuse reflection models are as follows:
\begin{equation}\label{eq2}
	I_s=F_s \cdot (\vec{N} \cdot \vec{H})^{\frac{1}{F_r}}
\end{equation}
\begin{equation}\label{eq3}
	I_d=F_d \cdot I_{sig} \cdot (\vec{N} \cdot \vec{L})^{F_b}
\end{equation}
For the specular reflection model $I_s$, where $F_s$ is the specular reflection coefficient; $F_r$ represents surface roughness; $\vec{N}$ is the surface normal vector; $\vec{H}$ is the bisector vector. For the diffuse reflection model $I_d$, where $F_d$ is the diffuse reflection coefficient; $I_{sig}$ represents the intensity of the incident signal; $\vec{L}$ is the normalized signal vector pointing from the surface point towards the SAR; $F_b$ is the surface brightness factor.

\begin{figure}[ht]
\vskip 0.2in
\begin{center}
\centerline{\includegraphics[width=0.8\columnwidth]{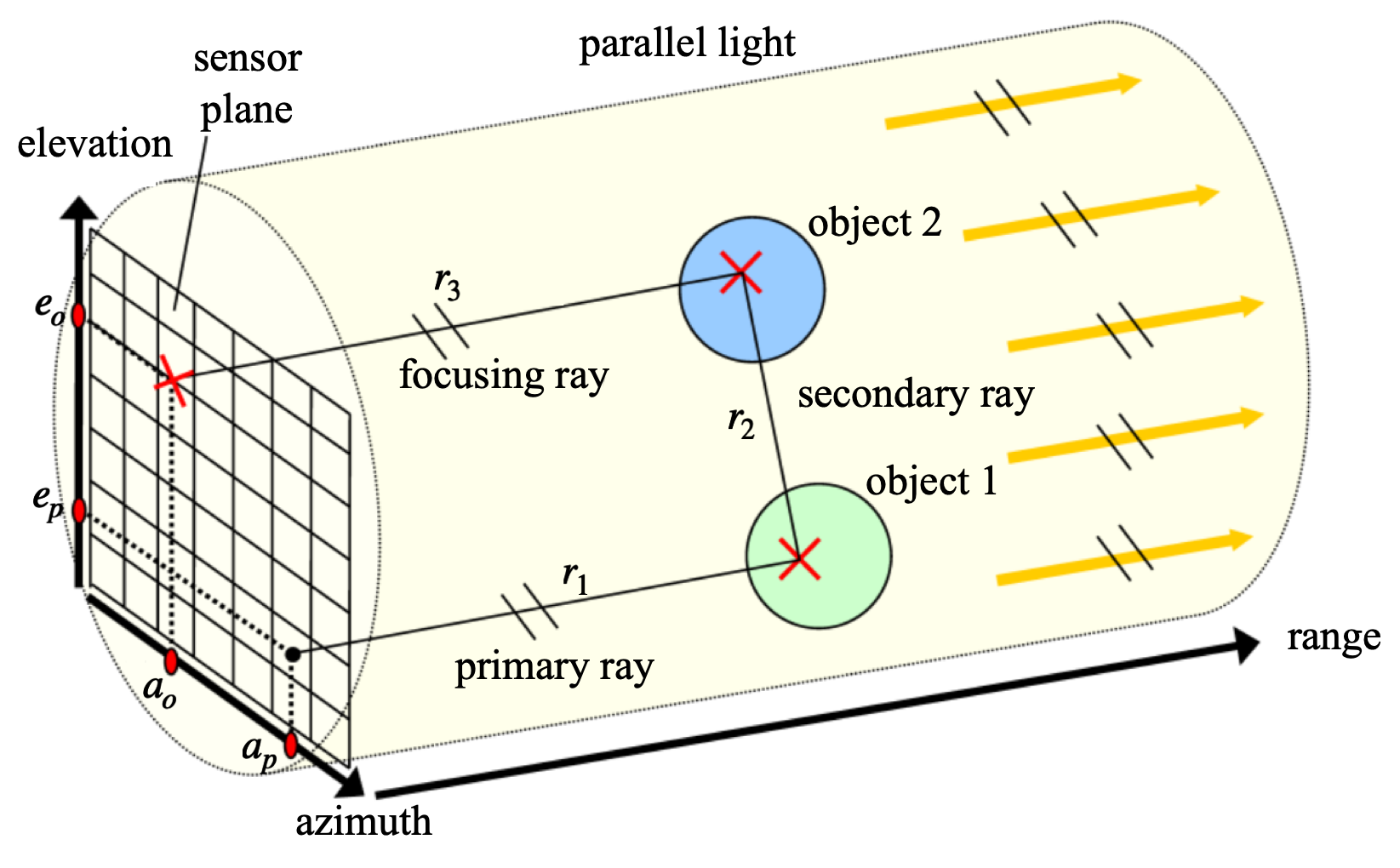}}
\caption{Schematic diagram of RaySAR imaging model.}
\label{figure3}
\end{center}
\vskip -0.2in
\end{figure}

In the imaging model, RaySAR represents the radar signal using parallel rays and performs orthogonal projection on the azimuth and elevation angles of the local scene. The position of the signal echo is calculated by projecting the starting point of the focusing ray onto the azimuth and elevation directions, respectively. As shown in Fig. 3, we use double reflection as an example to explain how to calculate the position of the signal echo, and the formula is expressed as follows:
\begin{equation}\label{eq4}
	a=\frac{x_o+x_p}{2}
\end{equation}
\begin{equation}\label{eq5}
	r=\frac{1}{2}(r_1+r_2+r_3)
\end{equation}
Where $a$ and $r$ are the coordinates in the azimuth and range directions, respectively; $x_o$ and $x_p$ are the coordinates of the main ray's incidence point and the focused ray's exit point projected onto the azimuth direction, respectively; $r$ is the sum of the lengths of rays emitted from and returning to the sensor plane after double reflection, which is $r_1+r_2+r_3$.

Our attack strategy involves altering the echo intensity of the target object by adjusting its adversarial scattering feature parameters, resulting in an adversarial example after imaging processing. Specifically, for the scattering feature parameter $t_i=\{F_{s-obj}^i,F_{d-obj}^i,F_{r-obj}^i,F_{b-obj}^i\}$ of each mesh composing the target object, we achieve the attack objective by adding perturbations to its reflection coefficient $F_{s-obj}^i$ and scattering coefficient $F_{d-obj}^i$. As indicated by formulas (2) and (3), a reduction in the reflection and scattering coefficients leads to a decrease in the target echo intensity. These changes further affect the grayscale characteristics of the SAR image. Particularly in the presence of background noise, they render the geometric features and scattering characteristics of the target object unclear, thereby making it difficult for the model to correctly identify or classify the target object.

To reduce the contrast between the target object and its surrounding environment in SAR images, we add a set of random coefficients $\alpha_i$, $\beta_i$, and $\gamma_i$ within the range of 0-1 to the surface parameter $t_i$ of each mesh of the target object. Then, we introduce perturbations into these random coefficients to generate adversarial scattering feature parameters $t_{i-adv}$. When perturbations are added to the random coefficients, it causes the reflection coefficient $F_{s-obj}^i$ and scattering coefficient $F_{d-obj}^i$ in the target object's scattering feature parameters to decrease. The purpose of this is to adjust the scattering feature parameters of the target object in SAR images, making it visually closer to the background and reducing the likelihood of being recognized by the classifier. The formula is expressed as follows:
\begin{equation}\label{eq6}
	\begin{cases}
 F_{s-obj-adv}^i=[1-(\alpha_i+\delta)] \cdot F_{s-obj}^i+(\alpha_i+\delta)\cdot F_{s-bg}^i \\
 F_{d-obj-adv}^i=[1-(\beta_i+\delta)] \cdot F_{d-obj}^i+(\beta_i+\delta)\cdot F_{d-bg}^i
 \end{cases}
\end{equation}

Wherein, $F_{s-bg}^i$ and $F_{d-bg}^i$ respectively represent the reflection coefficient and scattering coefficient of the surrounding environment. $\delta$ represents the size of the added perturbation, while $F_{s-obj-adv}^i$ and $ F_{d-obj-adv}^i$ represent the adversarial reflection coefficient and adversarial scattering coefficient of the target object, respectively.

\subsection{Scattering Feature Parameters Optimization}

After obtaining SAR adversarial example images through simulation experiments, we need to continuously iterate and optimize attribute parameters such as the reflection coefficient and scattering coefficient in the scattering feature parameters. This process aims to find a set of parameters that result in the highest misclassification rate for a given category by the classifier. Due to the non-differentiable nature of the RaySAR simulator, we estimate gradients effectively through differential approximation and utilizes batch processing techniques to achieve precise and efficient optimization of scattering feature parameters.

Specifically, first, the classifier operates in a no-gradient mode to ensure that no weight updates occur during the inference process. For each batch of images, we use the classifier to predict the category of each image and calculate the cross-entropy loss between the predictions and the actual labels. The loss calculation involves iterating over the entire dataset, accumulating the total loss for all images, and dividing by the total number of images to calculate the average loss. This average loss is used in the subsequent optimization process to evaluate the effectiveness of the current scattering feature parameter configuration. Gradient calculation is based on this, where we employ the finite difference method, as shown in Formula (7-10). For each scattering feature parameter $t_i$, we first apply a small positive perturbation $\delta$ to it and then recalculate the loss. The gradient is estimated by the difference in loss under the perturbed parameter versus the original parameter, divided by the perturbation amount $\delta$. Thus, for each scattering feature parameter, we obtain a gradient value indicating the extent of its impact on the final loss. The formula is expressed as follows:
\begin{equation}\label{eq7}
        loss_{now} = J(f(\Re(M,t_{i-adv-now};\theta_r);\theta_f),Y)
\end{equation}
\begin{equation}\label{eq8}
        loss_{before} = J(f(\Re(M,t_{i-adv-before};\theta_r);\theta_f),Y)
\end{equation}
\begin{equation}\label{eq9}
	\nabla t_{i-adv} = \frac{loss_{now}- loss_{before}}{\delta}
\end{equation}
\begin{equation}\label{eq10}
        t_{i-adv-now} = t_{i-adv-before}+\delta
\end{equation}
In the actual computation process, since the target model is composed of tens of thousands or even hundreds of thousands of meshes, recalculating the gradient for the scattering feature parameter $t_i$ of each mesh and updating it in reverse individually would greatly waste computational resources. To save computational overhead and accelerate the divergence speed of the loss function, we apply the method of mini-batch gradient ascent for optimization. Specifically, we treat the model's scattering feature parameters as a set of optimizable variables and divide them into several mini-batches $T=\{T_1,T_2,T_3,...,T_m\}$. In each iteration, we select a batch of parameters $T_i=\{t_1,t_2,t_3,...,t_{n/m}\}$ for optimization, rather than updating all parameters at once. This way, we can effectively approximate the global optimum while maintaining relatively low memory usage and computational complexity.

The experiments were conducted using the Adam optimizer to execute the optimization process. In each batch, by applying minor perturbations to the selected parameters and observing changes in the loss function, we estimated the gradients of these parameters. Then, utilizing these estimated gradient values, combined with the adaptive learning rate feature of the Adam optimizer, we updated these parameters. To further enhance the efficiency and stability of the optimization process, we also introduced a learning rate scheduler, adjusting the learning rate dynamically to accommodate the trend of changes in the loss function. When the average loss reaches a preset threshold, the learning rate drops by an order of magnitude, enabling more precise model parameter adjustments. Moreover, we employed gradient clipping techniques to prevent gradient explosion issues during the optimization process. Gradient clipping, by limiting the maximum norm of the gradient, ensures that the step size of gradient updates is not too large, thereby ensuring the stability of the optimization process. The formula is expressed as follows:
\begin{equation}\label{eq11}
	\nabla T_{i-adv}=Clip(\nabla T_{i-adv},-\epsilon,\epsilon)
\end{equation}
\begin{equation}\label{eq12}
	T_{i-adv}=T_{i-adv}+\eta \cdot Adam(\nabla T_{i-adv})
\end{equation}
Wherein, $\nabla T_{i-adv}$ represents the gradient value of a batch of adversarial scattering feature parameters; $Clip(\nabla T_{i-adv},-\epsilon,\epsilon)$ indicates limiting the gradient value within a range $[-\epsilon,\epsilon]$ using gradient clipping techniques; $\eta$ denotes the learning rate; $Adam(\nabla T_{i-adv})$ refers to updating the batch of adversarial scattering feature parameters $T_{i-adv}$ using the Adam optimizer.

\section{Experiments and Analysis}

\subsection{Experimental Dataset}

The MSTAR database, serving as a universal library for SAR image automatic target recognition research, is widely applied in adversarial attack studies. This paper conducted simulation experiments on target categories found in the MSTAR dataset. The experiments used point light sources, maintaining imaging parameters such as azimuth angle, elevation angle, and resolution consistent with real images. The dataset was generated under simulation conditions with an elevation angle of 15º and azimuth angle intervals of 1º, and divided into training and test sets in a 7:3 ratio. Then, we trained the simulation dataset. Due to the small amount of experimental data, the model was able to fit the training data well. After 20 epochs of iteration, the trained model achieved 100\% recognition accuracy on the test set.

\subsection{Experimental Setup}

We conducted detailed attack experiments on a simulation dataset to evaluate the proposed SAR-AE-SFP-Attack method. The experiments were divided into three parts: (1) Performance experiments: to verify the attack performance of the SAR-AE-SFP-Attack method across different classification models and target categories, and to compare SAR-AE-SFP-Attack with general adversarial perturbations and adversarial patch attacks currently studied in the SAR community; (2) Hyperparameter experiments: to test the impact of hyperparameters such as iteration numbers and target compositional structures on the attack performance of adversarial examples; (3) Transferability experiments, to test the transferability of the SAR-AE-SFP-Attack model against changes in viewpoints and models.

\subsection{Performance Experiments}

We first evaluated the attack effectiveness of the SAR-AE-SFP-Attack method on the T72 target category within the simulated dataset. In the experimental parameter settings, we set the size of each batch to 1/20 of the total number of meshes of the target, and the experiment iterated for a total of 25 epochs. The initial learning rate and the size of perturbations were all set to 0.001. Then, adversarial examples were generated at 10º intervals of azimuth angle under simulated conditions with an elevation angle of 15º. We compared the attack effectiveness of the SAR-AE-SFP-Attack method with Universal Adversarial Perturbations (UAP), adversarial patches (Patch), and the addition of random perturbations (Random) to the scattering feature parameters of the 3D target model in the T72 category. The experimental results are shown in Fig. 4.

\begin{figure}[ht]
\begin{center}
\centerline{\includegraphics[width=\columnwidth]{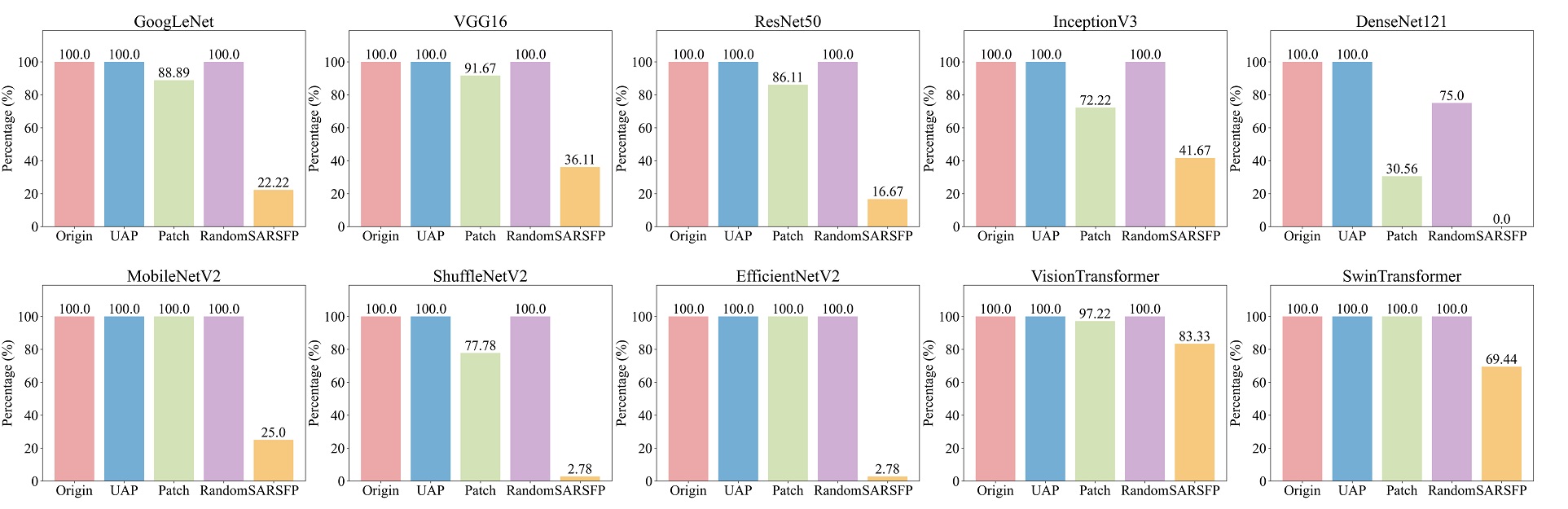}}
\caption{Comparison of attack effectiveness across different classifiers.}
\label{figure4}
\end{center}
\vskip -0.4in
\end{figure}

\begin{table}[t]
\caption{Comparison of accuracy before and after attack across different target categories.}
\label{table1}
\begin{center}
\setlength{\tabcolsep}{4pt}
\begin{scriptsize}
\begin{sc}
\begin{tabular}{lccccc}
\toprule
Model & Origin & UAP & Patch & Random & \makecell{SAR-AE-SFP-Attack} \\
\midrule
2S1      & 100.0\% &97.22\%  & 100.0\% & 88.89\% & \textbf{2.78\%} \\
BRDM-2   & 100.0\% & 100.0\% & 100.0\% & 100.0\% & \textbf{0.00\%} \\
BTR-60   & 100.0\% & 100.0\% & 100.0\% & 100.0\% & 100.0\% \\
T72      & 100.0\% & 86.11\% & 100.0\% & 100.0\% & \textbf{16.67\%} \\
ZIL131   & 100.0\% & 91.67\% & 100.0\% & 100.0\% & \textbf{13.89\%} \\
ZSU-23-4 & 100.0\% & 97.22\% & 100.0\% & 100.0\% & \textbf{80.56\%} \\
\bottomrule
\end{tabular}
\end{sc}
\end{scriptsize}
\end{center}
\vskip -0.2in
\end{table}

From Fig. 4, it can be observed that the trained classifier model slightly overfits the simulated data for the T72 category, causing the UAP method to lose its attack effectiveness. The Random method only demonstrated attack capability in DenseNet121, resulting in a 25\% decrease in the model's recognition accuracy for the T72 category. The Patch-based attack method was able to cause varying degrees of accuracy reduction in some classifier models. However, compared to the method presented in this paper, its attack effectiveness was still not significant. In contrast, the SAR-AE-SFP-Attack method improved the attack effectiveness by more than 30\% in classifier models based on the CNN architecture; for models based on the Transformer architecture, the attack effectiveness increased by more than 13\%.

Furthermore, we observed an interesting phenomenon: CNN models are easier to attack compared to Transformer models. Within the same architecture of network models, the more complex the model structure, the better the effectiveness of the SAR-AE-SFP-Attack method. The speculated reason is that complex networks overfit the background speckle noise present in SAR images. When the scattering feature parameters of the target object change, it causes a change in the contrast between the foreground and background of the imaging results, thereby making the attack more pronounced. The visual effects of different attack methods are shown in Fig. 5.

\begin{figure}[ht]
\begin{center}
\centerline{\includegraphics[width=\columnwidth]{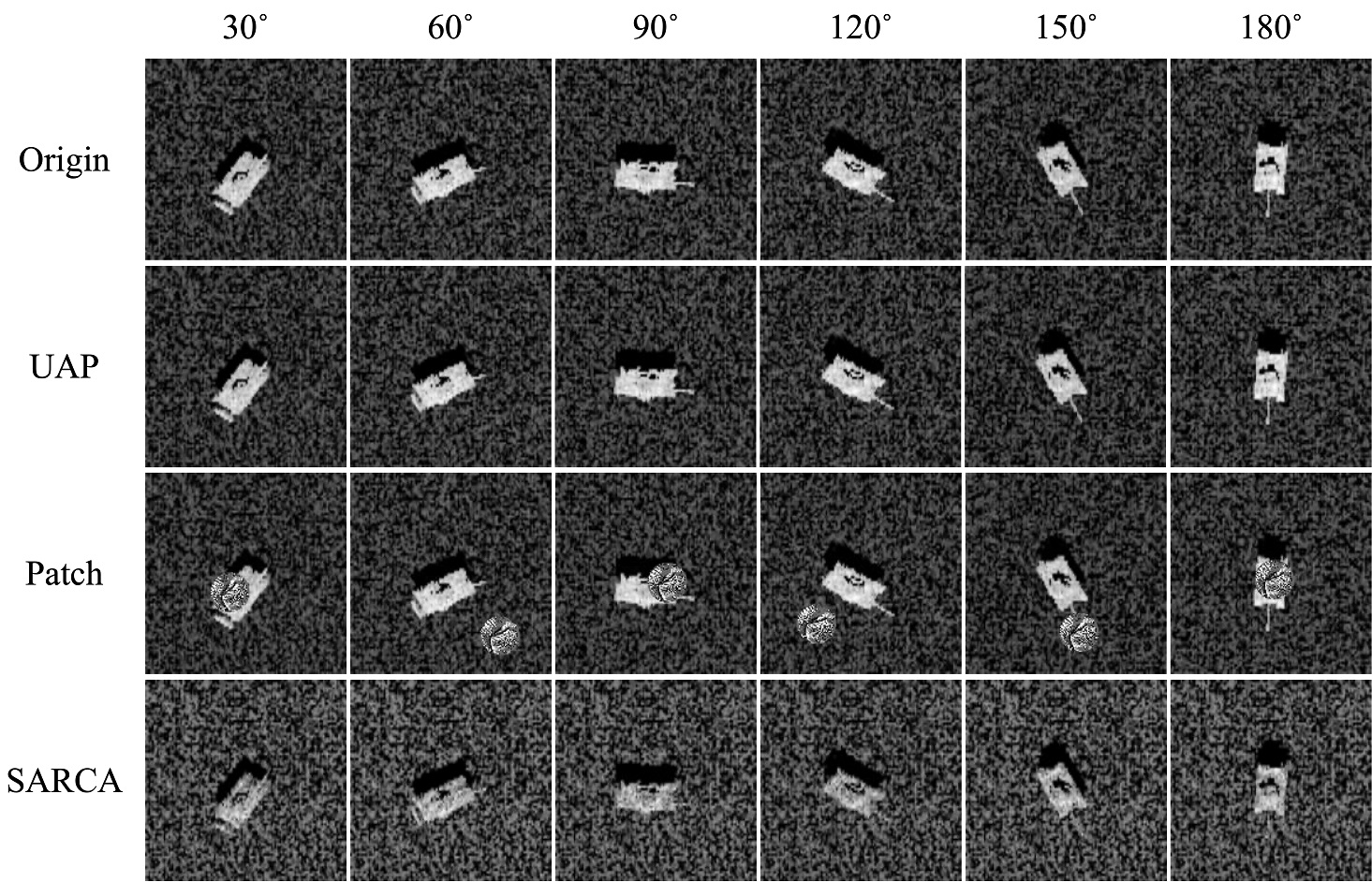}}
\caption{Visual comparison of different attack methods (ResNet50 model).}
\label{figure5}
\end{center}
\vskip -0.4in
\end{figure}

In the subsequent experiments, we evaluated the attack effectiveness of several attack methods on the ResNet50 model against different target categories. The experimental results are shown in Table 1. From Table 1, it can be observed that compared to other attack methods, our proposed SAR-AE-SFP-Attack method demonstrates strong attack capability, improving the attack effectiveness by more than 70\% across the majority of target categories compared to other methods. However, in the BTR-60 target category, all attack methods mentioned in this paper failed. The speculated reason is that the BTR-60 armored vehicle has significant differences in target contour and texture information compared to other categories in the dataset, making it difficult to be mistakenly recognized as other categories.

\subsection{Hyperparameter Experiments}

After validating the effectiveness of the SAR-AE-SFP-Attack method, we tested the impact of algorithm iteration numbers and target compositional structure on the attack performance for the T72 target category in the comparative experiment section. In the following experiments, we used the attack success rate as the standard to evaluate attack performance. The attack success rate is defined as:
\begin{equation}\label{eq13}
	\text{Attack Success Rate} = \frac{n_{\text{fooled images}}}{n_{\text{attacked images}}} \times 100\%
\end{equation}
Herein, $\text{fooled images}$ represents the number of successfully attacked examples, and $\text{fooled images}$ represents the total number of attacked examples.

\subsubsection{Number of Iterations}

The first test was on the impact of the number of iterations of the attack algorithm on its effectiveness. In the experiment, we set the number of iterations of the algorithm to 12, 16, 25, and 50 epochs, to assess the impact of different iteration counts on the success rate of adversarial sample attacks. According to the experimental results in Table 2, it can be observed that the attack success rate generally shows an upward trend with the increase in iteration numbers. However, it's noteworthy that compared to iterating for 50 epochs, the attack success rate achieved by iterating for 25 epochs was not significantly different and was even higher for certain classifiers (such as GoogLeNet and Inception). Therefore, to effectively save computational resources and avoid unnecessary waste, in subsequent experiments, we uniformly set the iteration count of the algorithm to 25 epochs.

\begin{table}[t]
\caption{Attack success rate for different iterations (T72).}
\label{table2}
\begin{center}
\setlength{\tabcolsep}{4pt}
\begin{scriptsize}
\begin{sc}
\begin{tabular}{lcccc}
\toprule
Model & 12 & 16 & 25 & 50 \\
\midrule
GoogLeNet       & 38.89\% & 61.11\% & \textbf{77.78\%} & 72.23\% \\
VGG16           & 58.33\% & 52.78\% & \textbf{63.89\%} & \textbf{63.89\%} \\
ResNet50        & 52.78\% & 58.33\% & 83.33\% & \textbf{86.11\%} \\
InceptionV3     & 38.89\% & 38.89\% & \textbf{58.33\%} & 52.78\% \\
DenseNet121     & \textbf{100.0\%} & \textbf{100.0\%} & \textbf{100.0\%} & \textbf{100.0\%} \\
MobileNetV2     & 41.67\% & 52.78\% & 75.00\% & \textbf{86.11\%} \\
ShuffleNetV2    & 86.11\% & 91.67\% & \textbf{97.22\%} & \textbf{97.22\%} \\
EfficientNetV2  & 80.56\% & 88.89\% & 97.22\% & \textbf{100.0\%} \\
VisionTransformer & 5.56\% & 8.33\% & \textbf{16.67\%} & \textbf{16.67\%} \\
SwinTransformer & 11.11\% & 33.33\% & 30.56\% & \textbf{55.56\%} \\
\bottomrule
\end{tabular}
\end{sc}
\end{scriptsize}
\end{center}
\vskip -0.3in
\end{table}

\subsubsection{Target Structure}

We further investigated the impact of the target's compositional structure on the success rate of attacks. Fig. 6 shows the disassembled structure of the T72 model, including six main components: the body, head armor, side skirts, turret, barrel, and track. In the experiments, we focused on evaluating the specific impact of individually adjusting the scattering feature parameters of any one component on the attack effectiveness. The related experimental data are detailed in Fig. 7. The results clearly indicate that the head armor plays a key role in enhancing attack effectiveness within the T72 target model. In contrast, the contributions of the body and barrel parts to attack effectiveness are relatively limited. As for the other components of the model, they only slightly contribute to enhancing the attack effect. The speculated reason is that the head armor, body, and barrel parts occupy most of the target surface area. When the scattering feature parameters of these parts change, it leads to alterations in the geometric characteristics and scattering properties of the target image, resulting in a higher success rate of attacks.

\begin{figure}[ht]
\begin{center}
\centerline{\includegraphics[width=\columnwidth]{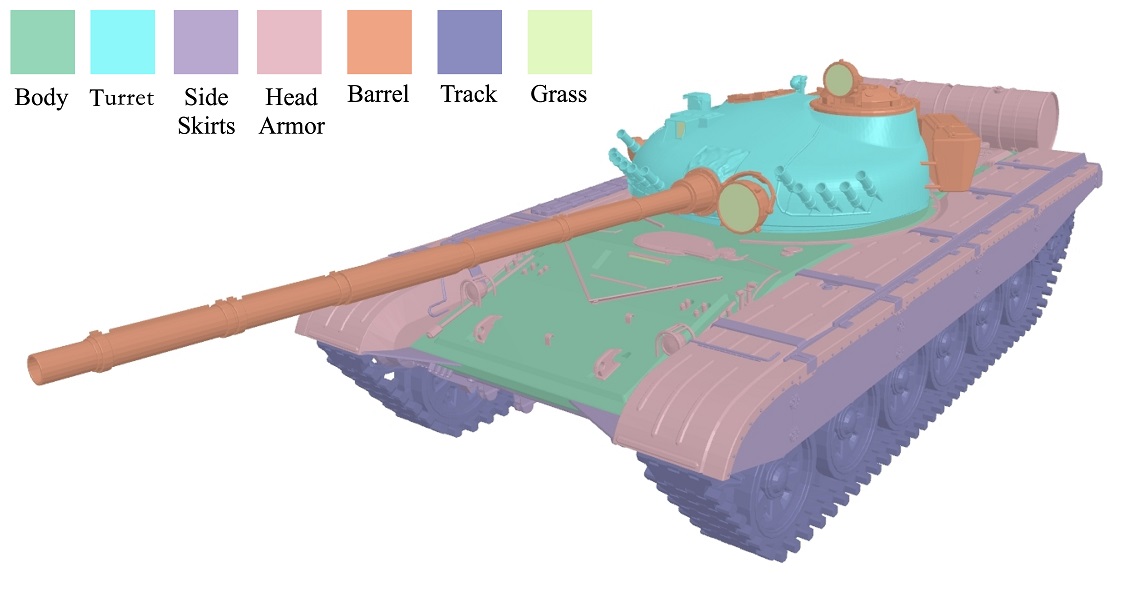}}
\caption{T72 target model breakdown structure diagram.}
\label{figure6}
\end{center}
\vskip -0.4in
\end{figure}

\begin{figure}[ht]
\vskip -0.1in
\begin{center}
\centerline{\includegraphics[width=\columnwidth]{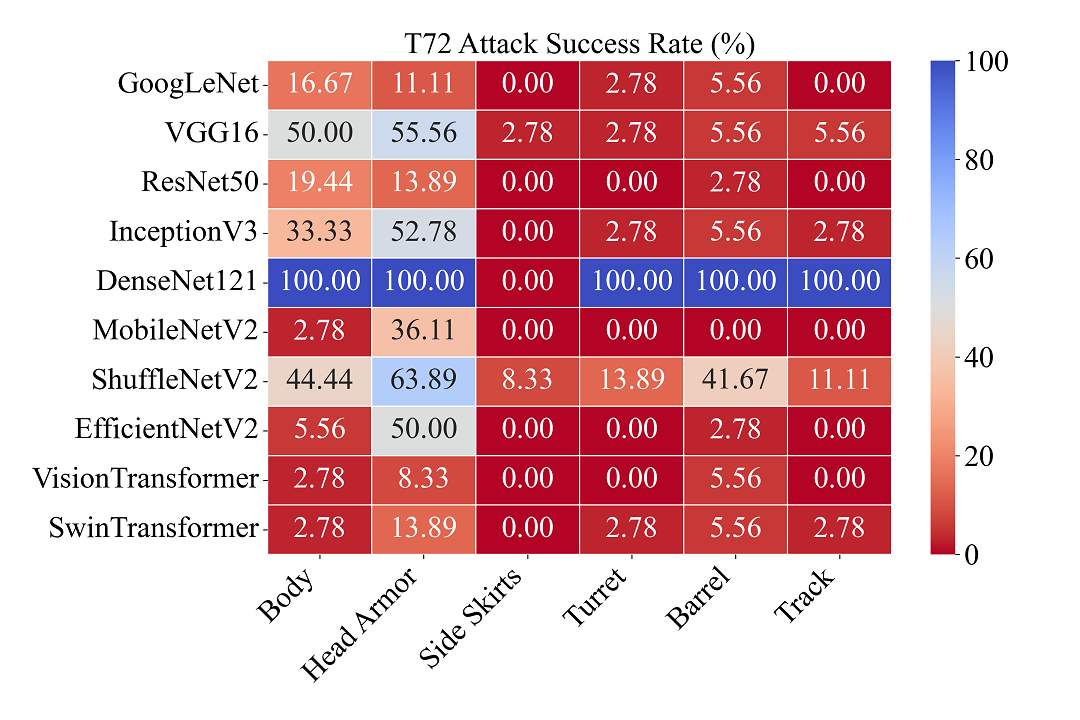}}
\caption{Impact of different compositional structures on the attack success rate of the T72 target model.}
\label{figure7}
\end{center}
\vskip -0.4in
\end{figure}

\subsection{Transfer Experiments}

In the final stage of the experiment, we tested the dual transferability of the SAR-AE-SFP-Attack method across different viewpoints and models. In previous experiments, we generated adversarial examples at an elevation angle of 15º with azimuth angle intervals of 10º under simulated conditions, and obtained the final adversarial scattering feature parameters $T_{adv}^*$ through iterative optimization. In the cross-viewpoint transfer experiment, we applied parameter $T_{adv}^*$ to generate adversarial examples at an elevation angle of 15º with azimuth angle intervals of 1º. A total of 360 adversarial SAR examples were obtained at different angles. Then, we divided the adversarial examples into 6 groups and tested the attack success rate of the adversarial examples on different classification models under 6 groups of azimuth angles. The experimental results are summarized in Fig. 8.

The experimental results show that SAR adversarial examples exhibit significant anisotropy, with varying attack effectiveness at different azimuth angles. Specifically, the most pronounced attack effectiveness occurs within the azimuth angle interval of $0\sim59$º, achieving an average attack success rate of 84.17\% across ten different classification models; conversely, the weakest attack effectiveness is observed within the azimuth angle interval of $120\sim179$º, with an average attack success rate of only 48.00\% across all classification models. The possible reason for this phenomenon is that, in the simulation experiments, the camera position is placed closer to the upper half region ($0\sim90$º, $270\sim360$º) for imaging, making the classifier more sensitive to changes in imaging parameters of images in the upper half region, thereby resulting in a higher success rate for attacks in that part.

\begin{figure}[ht]
\vskip -0.1in
\begin{center}
\centerline{\includegraphics[width=\columnwidth]{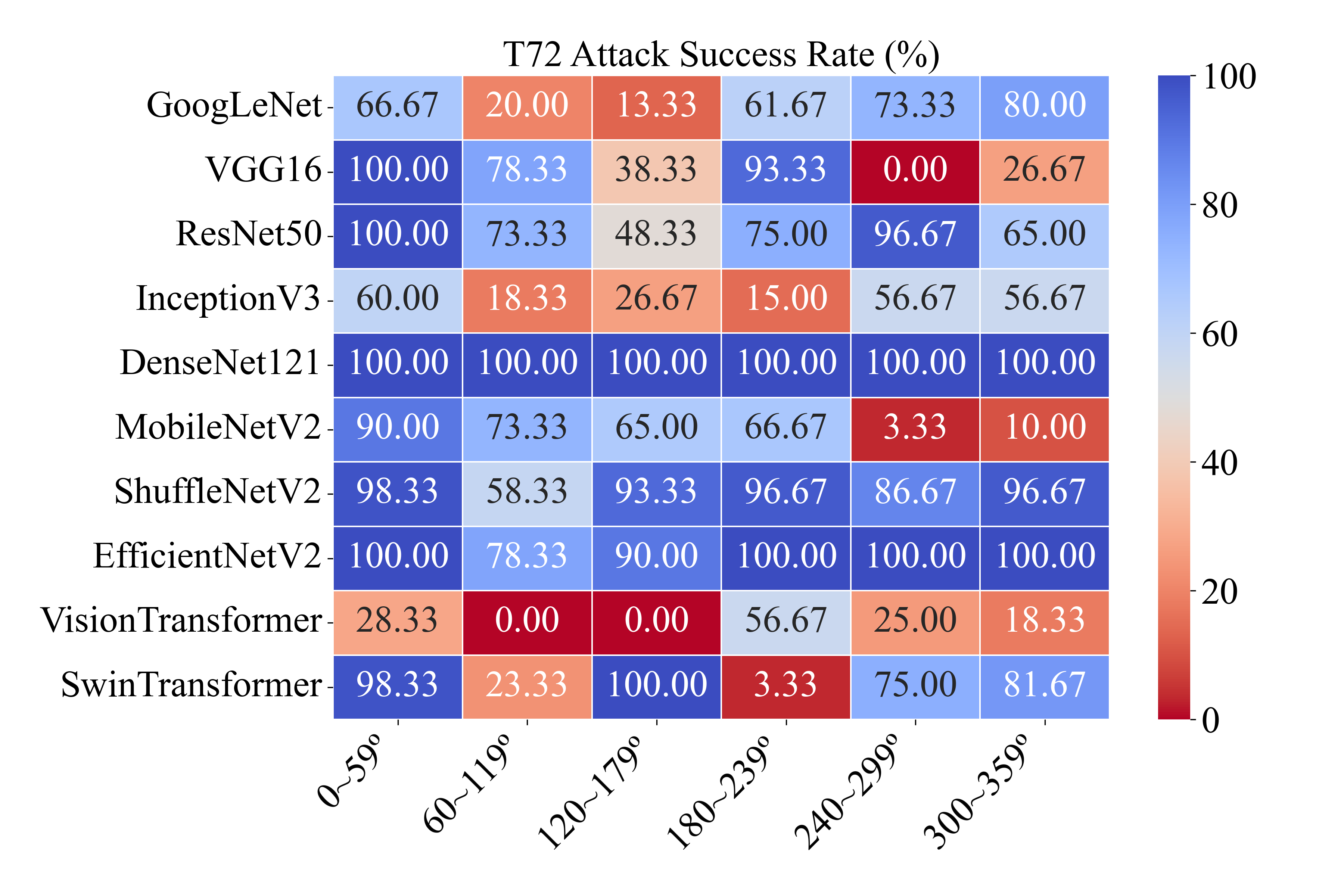}}
\caption{The impact of different azimuth ranges on attack success rate (T72).}
\label{figure8}
\end{center}
\vskip -0.4in
\end{figure}

After completing the cross-viewpoint transfer experiments, we further explored the transferability of SAR adversarial examples across different models. The experimental results shown in Fig. 9 reveal an interesting phenomenon: adversarial examples generated for a specific classification model not only retained their attack efficacy when transferred to other models, but in some cases, even surpassed their performance on the original model. This finding indirectly supports our previous view that, due to complex networks' tendency to overfit the background speckle noise in SAR images, changes in the scattering feature parameters of the target object cause changes in the contrast between the foreground and background of the imaging results. This, in turn, makes the attack effects more pronounced.

\begin{figure}[ht]
\vskip -0.1in
\begin{center}
\centerline{\includegraphics[width=1.1\columnwidth]{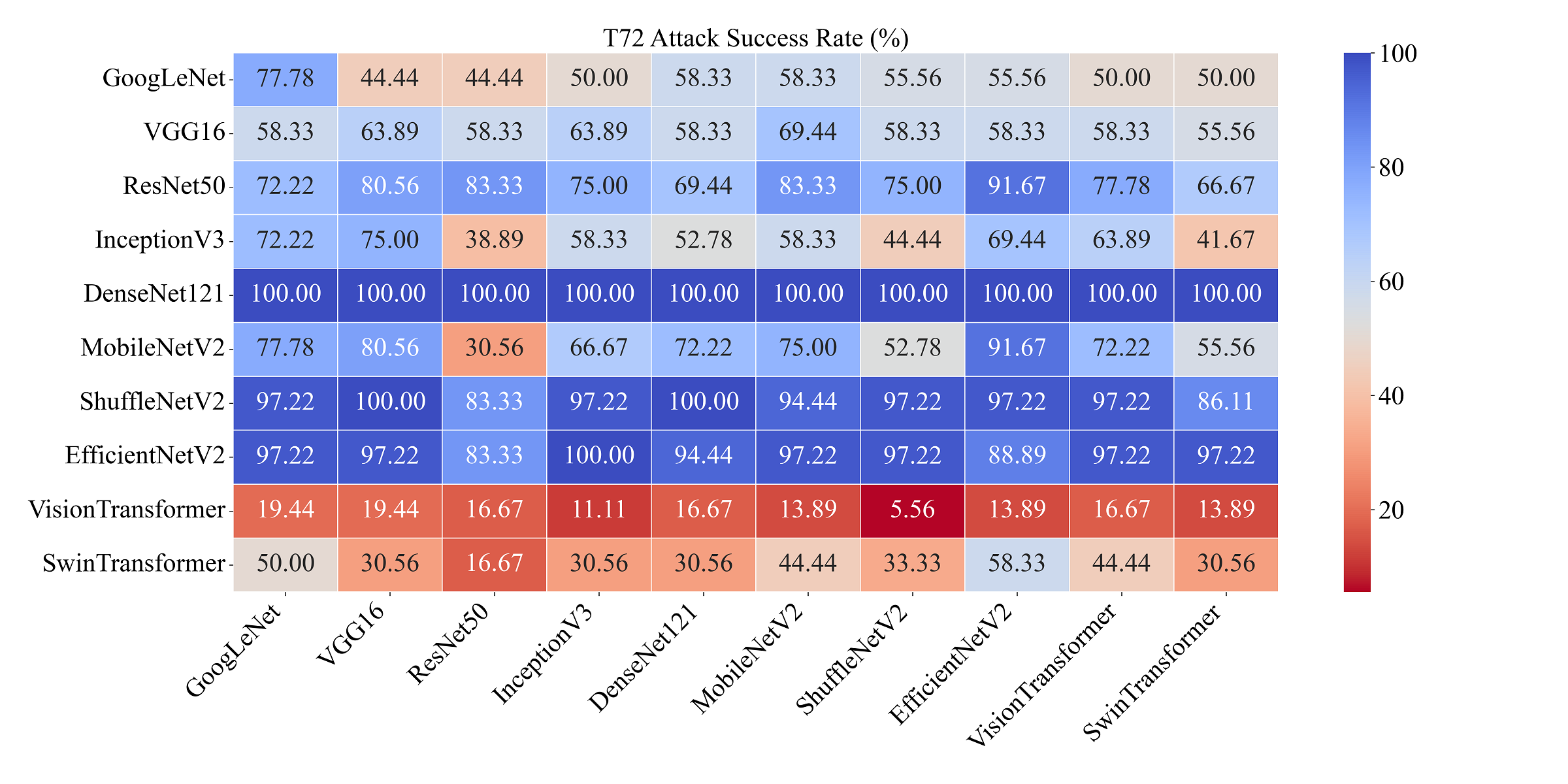}}
\caption{Heatmap of attack success rate for cross-model transfer experiments.}
\label{figure9}
\end{center}
\vskip -0.4in
\end{figure}

\section{Conclusion}

In this paper, we propose an end-to-end attack method, SAR-AE-SFP-Attack, for generating SAR image adversarial examples in the physical world. Initially, this method integrates SAR imaging mechanisms into the adversarial example generation process. By using the RaySAR physical simulator to adjust the scattering feature parameters of the target object, it changes the target echo intensity and obtains adversarial examples after imaging processing. Subsequently, we propose an optimization approach using differential substitution for differentiation. It effectively estimates gradients by making minor perturbations to the target's scattering feature parameters and utilizes batch processing techniques for precise and efficient optimization of adversarial scattering feature parameters, addressing the non-differentiability of the RaySAR simulator. Extensive experiments show that our SAR-AE-SFP-Attack method demonstrates higher attack effectiveness compared to other general attack methods and possesses transferability across various models and angles. Therefore, the SAR-AE-SFP-Attack method demonstrates potential for application in the physical world, offering an interesting direction for future work in physical attacks.


\bibliography{example_paper}
\bibliographystyle{icml2024}

\end{document}